\documentclass[letterpaper, 10 pt, conference]{ieeeconf}  % Comment this line out if you need a4paper

\IEEEoverridecommandlockouts                              % This command is only needed if 
\usepackage{graphicx} % for pdf, bitmapped graphics files
\usepackage{amsmath} % assumes amsmath package installed
\usepackage{amssymb}  % assumes amsmath package installed
\usepackage[caption=false]{subfig}
\usepackage{multicol}
\usepackage{booktabs}
\usepackage{color,soul}
\usepackage[ruled,vlined]{algorithm2e}
\usepackage{hyperref}
\SetKwInput{KwInput}{Input}
\SetKwInput{KwOutputput}{Output}

\title{\LARGE \bf
LRGNet: Learnable Region Growing for Class-Agnostic Point Cloud  Segmentation
}

\author{Jingdao Chen$^{1}$ , Zsolt Kira$^{1}$and Yong K. Cho$^{1}$% <-this % stops a space
\thanks{$^{1}$ Institute for Robotics and Intelligent Machines, Georgia Institute of Technology, Atlanta, GA 30332, USA
        {\tt\small jchen490@gatech.edu, zkira@gatech.edu, yong.cho@ce.gatech.edu}}
}

\begin{document}

\maketitle
\thispagestyle{empty}
\pagestyle{empty}

%%%%%%%%%%%%%%%%%%%%%%%%%%%%%%%%%%%%%%%%%%%%%%%%%%%%%%%%%%%%%%%%%%%%%%%%%%%%%%%%
\begin{abstract}
3D point cloud segmentation is an important function that helps robots understand the layout of their surrounding environment and perform tasks such as grasping objects, avoiding obstacles, and finding landmarks. Current segmentation methods are mostly class-specific, many of which are tuned to work with specific object categories and may not be generalizable to different types of scenes. This research proposes a learnable region growing method for class-agnostic point cloud segmentation, specifically for the task of instance label prediction. The proposed method is able to segment any class of objects using a single deep neural network without any assumptions about their shapes and sizes. The deep neural network is trained to predict how to add or remove points from a point cloud region to morph it into incrementally more complete regions of an object instance. Segmentation results on the S3DIS and ScanNet datasets show that the proposed method outperforms competing methods by 1\%-9\% on 6 different evaluation metrics.
\end{abstract}

%%%%%%%%%%%%%%%%%%%%%%%%%%%%%%%%%%%%%%%%%%%%%%%%%%%%%%%%%%%%%%%%%%%%%%%%%%%%%%%%
\section{INTRODUCTION}

A key step in parsing a 3D point cloud scene is point cloud segmentation. Segmentation is useful for many robotic applications such as obstacle detection \cite{chen18ur}, localization \cite{dube18}, and object recognition \cite{kim17}\cite{qi17}. In this research, point cloud segmentation is defined as subdividing a point cloud scene into individual object segments where each point in the point cloud is assigned to a unique instance label (i.e. class-agnostic segmentation). One popular type of approach is to perform region growing to iteratively create different point cloud segments \cite{law18}\cite{rusu10}\cite{wang2017}. However, this approach is highly sensitive to specific threshold hyperparameters defined over the raw feature values. Another popular type of approach is to directly learn the segmentation labels using a deep neural network \cite{wangweiyue18}\cite{pham19}\cite{yang19}. These approaches are data-driven and are usually more accurate and robust, but require discretizing the point cloud scene into 1m x 1m blocks \cite{qi17}\cite{wangweiyue18} and need class annotations during training. We propose to extend these two types of approaches, combining the region growing approach with data-driven deep learning techniques in order to achieve the benefits of both types of approaches.

Thus, this research proposes a point cloud segmentation method where the region growing process itself is learned from data. Given a raw point cloud scene, the proposed method tries to form point cloud segments that each correspond to a complete object by making a series of incremental predictions. Specifically, a deep neural network is trained to predict at each decision point: (i) which points to add to grow the current region, (ii) which unnecessary points to remove from the current region, conditioned on the current set of points. The proposed method is evaluated on the S3DIS \cite{armeni16} and ScanNet \cite{dai17} datasets with a focus on generalization performance across datasets. The proposed method has the following advantages: 
\begin{itemize}
    \item able to segment any class of objects using a single neural network
    \item able to segment objects without assumptions about their shapes and sizes
    \item can be trained without semantic labels
    \item does not require block division of the point cloud scene
    \item more accurate and generalizable compared to existing methods
\end{itemize}
In addition, the code and data have been made publicly available. \footnote{\url{https://github.com/jingdao/learn\_region\_grow}} 

\section{LITERATURE REVIEW}

Point cloud segmentation can take the form of:  (i) semantic label prediction -  the task of predicting a class label (e.g. wall, window) for each point of a given point cloud scene, or (ii) instance label prediction, the task of assigning points to unique objects instances in a given point cloud scene. This research will specifically focus instance label prediction, which is especially challenging in real-world point cloud scenes because they are often cluttered and contain objects with a wide variety of shapes and sizes.

\subsection{Class-agnostic segmentation}

Class-agnostic methods generally perform segmentation without making any assumptions about the class of objects that are in the point cloud scene. Class-agnostic methods are generally instance-based segmentation methods. In the robotics domain, these methods are commonly implemented as region growing, where groups of points with similar features such as surface normal and color are incrementally merged together \cite{dube18}\cite{rusu10}. These methods face issues with oversegmentation or undersegmentation in cluttered scenes due to the reliance on fixed thresholds or heuristics to control the region growing process. Some works attempt to use reinforcement learning \cite{sahba06}\cite{bhardwaj17} or evolutionary methods \cite{derivaux10} to more effectively tune the threshold hyperparameters, but do not directly optimize the region growing process itself. An alternative to region growing is to perform segmentation as a global operation by using clustering algorithms such as K-means \cite{zhu14} or DBSCAN \cite{ahmed20}. The problem with these clustering algorithms is that the results are sensitive to the hyperparameters such as number of segments and the neighborhood range. Other ideas for class-agnostic segmentation include using a network to predict object seeds \cite{xie20} or searching over the space of candidate segmentations and returning one which scores well according to an “objectness” model \cite{hu20}. However, the former method was only evaluated for the tabletop setting where most objects are roughly the same size whereas the latter method was only applied in outdoor environments where objects do not physically overlap with one another.

\subsection{Class-specific segmentation}

Class-specific segmentation methods try to separate objects from the background based on prior information about the geometry of particular classes of objects. Class-specific methods are generally semantic segmentation methods. These methods are usually data-driven and make use of deep neural networks. A popular approach is to use 3D bounding box regression to predict the location and size of objects in the scene, used by methods such as MV3D \cite{mv3d}, PointRCNN \cite{yang19}, PointPillars \cite{lang19}, and 3D-BoNet \cite{shi19}. The downside of this approach is the need to define class-specific anchor boxes \cite{mv3d} or train multiple networks separately for each class \cite{liu17}, which is not scalable for a large number of classes. In addition, bounding boxes only provide a coarse outline and do not cleanly separate object segments. A different strategy is to directly predict labels for each point in the point cloud using neural networks based on PointNet \cite{qi17} or PointNet++ \cite{qi17_plus}. These network are usually used to perform semantic segmentation or part segmentation and do not directly predict the instance labels. Later methods combine the semantic segmentation results with grouping matrix prediction \cite{wangweiyue18}\cite{jianghaiyong20}, centroid prediction \cite{jiang20}\cite{engelmann20}, or clustering on a learnt embedding \cite{pham19}\cite{he20} in order to obtain instance segmentation results. However, a common problem is that they require a pre-processing step of subdividing the point cloud into fixed-size blocks and a post-processing step of \textit{BlockMerging} \cite{wangweiyue18} to combine segments that span different blocks together, which leads to information loss. Another weakness is that they rely on having training data that is labeled with class information and ignore objects that are not within the predefined set of classes (usually designated as "clutter"). In contrast to the proposed method which is class-agnostic, these class-specific segmentation methods are difficult to generalize across datasets which have different classes.

\section{METHODOLOGY}

This section describes the proposed LRGNet method for class-agnostic point cloud segmentation including the formal definition, network architecture, training process, inference time algorithm, and local search optimization.

\begin{figure*}
  \centering
  \includegraphics[width=0.99\linewidth]{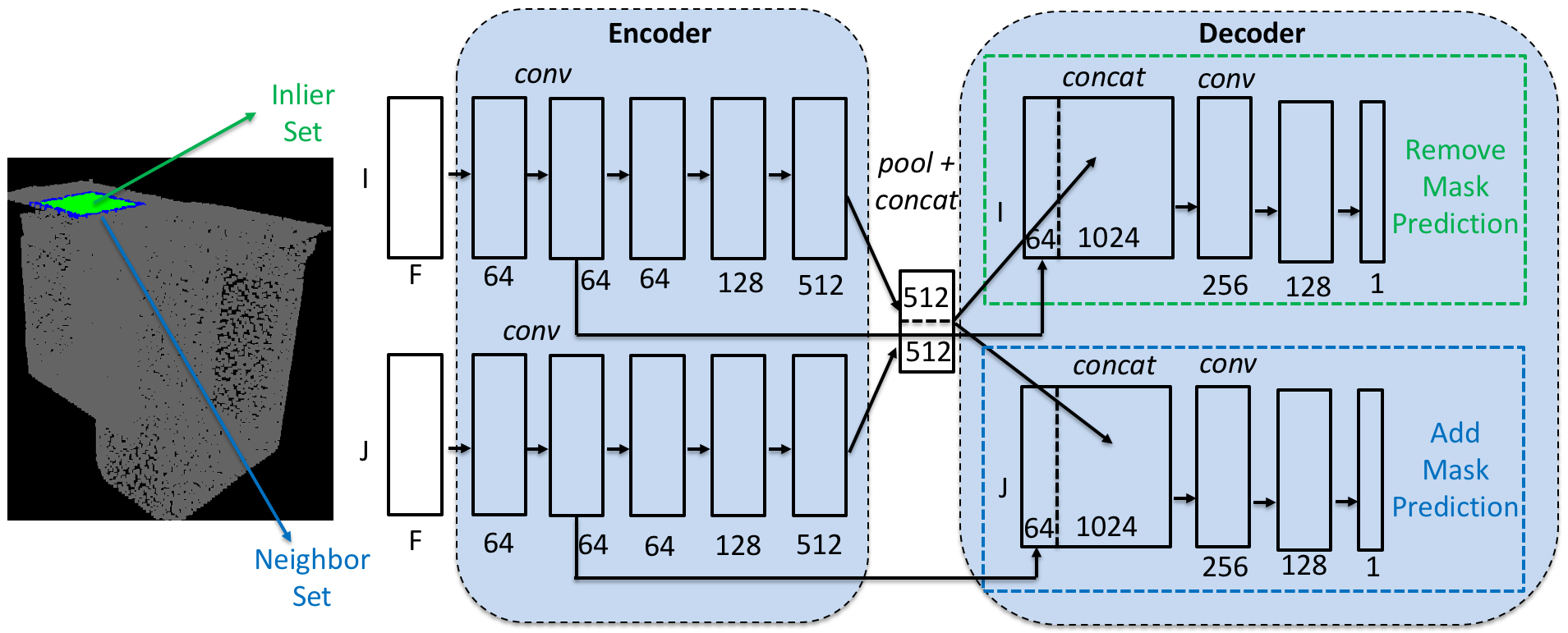}
  \caption{LRGNet network architecture to perform an intermediate step of region growing. The network takes as input two sets of points: the inlier set and the neighbor set. Features are computed from both sets of points and then pooled and concatenated to form a global feature vector. This feature vector is concatenated back into the point-level features to predict which points in the inlier set to remove (i.e. \textit{remove mask} prediction) and which points in the neighbor set to add (i.e. \textit{add mask prediction})}
  \label{fig:architecture}
\end{figure*}

\subsection{Formal definition}

The point cloud segmentation problem is defined as follows. The input is a point cloud $\textbf{P}$, represented as a $N \times F$ matrix of real numbers where the input point cloud contains $N$ points with each point having $F$ features (e.g. 3D position, color, etc.).  The task is to output the instance labels $\textbf{L}$, represented as a $N$-dimensional vector of integers, which contains one integer label for each point indicating which object instance it should belong to.

The segmentation problem can be decomposed into several region-growing subproblems. The region-growing procedure takes in an input seed point $p_{seed} \in \textbf{P}$, and outputs a set of points $P^* \subset P$, that all belong to the same object instance as $p_{seed}$. The instance label vector, $\textbf{L}$, is then updated such that all points in $P^*$ have the same instance label. The idea is that this region-growing procedure can be carried out repeatedly from any remaining unlabeled seed points in the point cloud $\textbf{P}$ until all points have been assigned an instance label.

In this study, the region-growing subproblem is further decomposed into a series of steps, where each step takes in an input set of points $Q_k$ and outputs a set of points $Q_{k+1}$, where $Q_k,Q_{k+1} \subset \textbf{P}$. This can be represented as $f(Q_k) = Q_{k+1}$, where $f$ is a learnable function. The region-growing process starts with $Q_0 = \{p_{seed}\}$ and computes $Q_1,Q_2,Q_3,...$ etc. until $Q \rightarrow P^*$, the set of all points which belong to the same object instance as $p_{seed}$. This study aims to learn the function $f$ which dictates how a point cloud region grows from one set of points to another. The key novelty of this formulation is to decompose the region growing problem and use a \textit{learnable function} to carry it out, thus achieving a "learnable" region growing method.

\subsection{Network architecture}

The learnable function $f(Q_k) = Q_{k+1}$ takes the form of a deep neural network named LRGNet, as shown in Figure \ref{fig:architecture}. The input to the network consists of two sets of points, the inlier set (represented as an $I \times F$ matrix) and the neighbor set (represented as a $J \times F$ matrix). The inlier set is constructed by taking $I$ points from $Q_k$. Note that the full set of points in the current region, $Q_k$, is maintained as a separate global variable. The neighbor set is constructed by taking all the unlabeled points that are within a predetermined $\delta$ distance of any point in $Q_k$ and selecting $J$ points. When there are more or fewer than $I$ or $J$ points in the subsets, random resampling with replacement is used to obtain exactly $I$ or $J$ points.

The network uses an encoder-decoder structure. First, the input inlier set and neighbor set are separately processed by individual network branches each containing 5 point-wise feature convolution layers. The features across all the points are then aggregated using a max-pooling layer to obtain a global feature vector incorporating 512 features from the inlier set and 512 features from the neighbor set. This global feature vector is then separately concatenated back into two decoder branches, one processing the inlier set and one processing the neighbor set. These two branches are then further processed by point-wise feature convolution layers. This network design is largely based off of PointNet \cite{qi17}, but with two input branches and two output branches instead of a single input branch and a single output branch.

The upper branch finally outputs an $I$-dimensional vector which constitutes the \textit{remove mask} probability prediction output. This is then used to determine the \textit{remove mask}, a binary mask which determines which points in the inlier set should be removed from $Q_k$. On the other hand, the lower branch finally outputs a $J$-dimensional vector which constitutes the \textit{add mask} probability prediction output. This is then used to determine the \textit{add mask}, a binary mask which determines which points in the neighbor set should be added to $Q_k$. The reason for having two separate masks is because the \textit{remove mask} prediction allows the network to correct for previous errors and makes it more robust to noise. Finally, the new set of points $Q_{k+1}$ is determined by applying the \textit{remove mask} to the inlier set and removing the corresponding points from $Q_k$, then applying the \textit{add mask} to the neighbor set and adding those points to $Q_k$.

This study uses a total of $F=13$ input features for each point, which are the local XYZ-coordinates, room-normalized XYZ-coordinates, RGB-color, normal vector, and curvature. The room-normalized XYZ-coordinates are obtained by dividing the X,Y,Z coordinates by the length, width, height of the room respectively to obtain values ranging from 0 to 1. The local XYZ-coordinates describe a point in relation to its object center whereas the room-normalized XYZ-coordinates describe a point in relation to the global room scene. The normal vector and curvature are computed using the Principal Component Analysis method from \cite{rusu10} whereas the XYZ features are computed as in \cite{qi17}. These point features (except the room-normalized XYZ-coordinates) are normalized column-by-column by subtracting the median over all the points in the inlier set. On the other hand, the parameters $\delta$ (neighbor threshold distance), $I$ (inlier set size), $J$ (neighbor set size) are set to 0.1m, 512, 512 respectively. These parameters are tuned based on an ablation study on the S3DIS dataset.

The network is trained on the binary cross entropy loss on both the add mask prediction and the remove mask prediction. This is shown in (1), where $\hat{x_i}$ denotes the predicted probability for removing point $i$ in the inlier set and $x_i$ is the corresponding ground truth label, whereas $\hat{y_j}$ denotes the predicted probability for adding point $j$ in the neighbor set and $y_j$ is the corresponding ground truth label.

\begin{equation}
\begin{split}
\mathcal{L} = -\frac{1}{I}\sum^I_i{[x_i \log{\hat{x_i}} + (1 - x_i) \log{(1 - \hat{x_i})}]} \\
-\frac{1}{J}\sum^J_j{[y_j \log{\hat{y_j}} + (1 - y_j) \log{(1 - \hat{y_j})}]}
\end{split}
\end{equation}

\begin{figure*}
  \centering
  \includegraphics[width=0.99\linewidth]{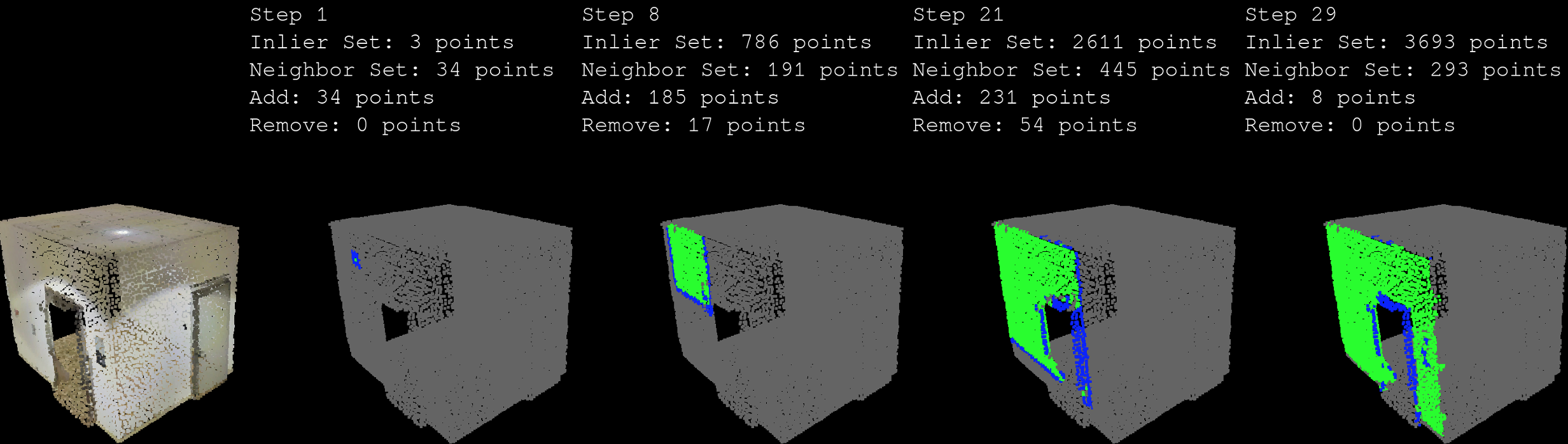}
  \caption{Visualization of four different intermediate steps of the region growing process starting from a seed point (points in the inlier set are shown in green whereas points in the neighbor set are shown in blue).}
  \label{fig:animation}
\end{figure*}

\subsection{Region growing simulation and training}

\begin{algorithm}[]
\caption{Region Growing}
\label{alg}
\KwInput{PointCloud, P}
\KwOutputput{InstanceLabels, L}
$\phi$ $\equiv$ empty label \;
$L = \{\phi, \forall p \in P\}$ \;
$obj\_id = 1$\;
\While{$\phi \notin L$}
{
    $p_{seed}$ = unlabeled point with min curvature\;
    $k = 0$\;
    $Q_k = \{p_{seed}\}$\;
    \While{true}
    {
        $inliers = sample(Q_k, count = I)$\;
        $neighbors = sample(\{p: |p-q|<\delta, p \in P, q \in Q_k, L[p]=\phi \}, count = J)$\;
        $add, remove = net(inliers, neighbors)$\;
        \eIf{termination condition}
        {
            $L[Q_k] = obj\_id$\;
            $obj\_id = obj\_id + 1$\;
            \textbf{break}\;
        } {
            $Q_{k+1} = Q_k - inliers[remove] + neighbors[add]$\;
            $k = k + 1$\;
        }
    }
}
\end{algorithm}

In this research, training data is obtained by carrying out a series of region growing simulations from labeled point cloud datasets, S3DIS \cite{armeni16} and ScanNet \cite{dai17}. The region growing process for each object instance in a point cloud scene is simulated in a series of steps, starting from a randomly selected seed point. At each step, the desired next point set, $Q_{k+1}$, is created from the current point set, $Q_k$, by adding all the points that are within $\delta$ distance of any point in $Q_k$ and also have the same ground truth instance label. To prevent the network from overfitting to only clean point cloud samples, noise is artificially introduced into the region growing process by randomly adding and removing points to $Q_{k+1}$ according to a mistake probability, $\alpha$. Within one region growing simulation, the mistake probability starts at a high value and is then gradually decreased by 0.01 each step for the current region to ensure that region growing process will eventually converge to a complete object instance.

Data augmentation is carried out in several ways. First, each point cloud scene is randomly flipped to swap the \textit{x} and \textit{y} axes. Next, the point cloud scene is randomly rotated in increments of 90$^{\circ}$. Also, the mistake probability, $\alpha$, for each object instance is randomly assigned an initial value between 0.2 and 0.4 to simulate different levels of noise in the region growing process. For example, after data augmentation on the S3DIS dataset, the training set consists of 1406516 point clouds originating from 63047 object instances.

The acquired training data is used to train LRGNet as described in Section IIIB, implemented in Tensorflow. Note that the training process only uses the ground truth instance labels and does not require semantic labels. The network is trained with the ADAM optimizer until convergence at around 40 epochs. The learning rate used is 0.001 and the batch size is 100.

\subsection{Region growing at inference time}

At inference time, segmentation of a point cloud scene is carried out by continuously performing region growing until each point in the scene is assigned an instance label. Starting from a seed point, the trained neural network is applied to predict the next region to grow to using the \textit{add mask} and \textit{remove mask} output.
This process is repeated until one of the following termination conditions is met:
\begin{itemize}
    \item No unassigned neighbor points remaining
    \item The set of points to be added is predicted to be empty
    \item Region does not expand for two consecutive steps
\end{itemize}
The last termination condition is necessary to avoid situations where different points are alternatingly added then removed on consecutive steps in an alternating manner, which could occur near edges. Upon reaching the termination condition, all the points in the current region areis assigned the same instance label and the region growing process starts over again from a new seed point. Note that the region points may change over the course of region growing for a single object instance but the assigned instance label will be permanent once the region growing process switches to a new object.

Since the region growing result could be variable due to the choice of the seed point, this study uses the strategy of picking the remaining point with minimum curvature as the seed point. This strategy is also used in \cite{dimitrov15}. In addition, as the region growing progresses, certain groups of points may become isolated. This frequently occurs for edges in between two object instances. To overcome this, if the resulting point cloud segment is too small (e.g. less than 10 points), then a new instance is not created. Instead, the points in that isolated segment are assigned instance labels according to the nearest neighbor among points that have already been assigned instance labels. A similar strategy is used in \cite{qi17}\cite{chen19jcce}.

Algorithm \ref{alg} shows a brief pseudocode summary of the region growing process for a given point cloud scene. To further illustrate the process, Figure \ref{fig:animation} shows a visualization of region growing for a wall object. As shown in Figure \ref{fig:animation}, there tends to be a higher number of added and removed points during the middle of the process and a lower number of added and removed points during the start and end of the process. On average, the percentage of added points and removed points per step are 70\% and 59\% respectively for a typical point cloud.

\subsection{Local search to optimize region growing}

Since the learned function $f(Q_k) = Q_{k+1}$ is not perfect and also depends on randomly sampling the inlier set and neighbor set, the region growing process may not converge on the perfect point cloud segment that contains all the points from one object instance. To overcome this, local search techniques can be applied to search through different sequences of applying the learned region growing function to find the best final region. This study considers 5 different methods of arriving at the final region from a given seed point: (i) \textit{greedy}: the \textit{add} and \textit{remove} outputs are greedily applied as binary masks with a cutoff threshold of 0.5, (ii) \textit{random restart - Maximum Likelihood (ML)}: the region growing process is restarted from the same seed point multiple times, each time probabilistically sampling points to be added and removed and keeping track of the sum of the log-likelihoods from each step; the best final region is determined using maximum likelihood as the optimality criterion, (iii) \textit{random restart - Number of Points (NP)}: same as the previous method but with the number of points in the final region as the optimality criterion (based on the assumption that regions with more points are more complete), (iv) \textit{beam search - ML}: multiple states for the current region are simultaneously tracked and expanded \cite{umbach94}; after each expansion step, the number of states is reduced to the $K$ best states using maximum likelihood as the optimality criterion, (v) \textit{beam search - NP}: same as the previous method but with the number of points as the optimality criterion. A comparison of the segmentation performance between different local search methods will be given in the Results section.

\section{RESULTS}

\subsection{Segmentation in indoor scenes}

\begin{table}[h]
  \caption{Comparison of segmentation performance on S3DIS and ScanNet by testing generalization across datasets}
  \label{table:acc}
  \centering
  \begin{tabular}{ccccccc}
    \toprule
    Method & NMI & AMI & ARI & PRC & RCL & mIOU\\
    \midrule
\multicolumn{7}{l}{\textit{ScanNet (train) $\rightarrow$ S3DIS (test)}} \\
Region growing & 0.71 & 0.70 & 0.59 & 0.19 & 0.34 & 0.38 \\
Rabbani et al. \cite{rabbani06} & 0.72 & 0.71 & 0.62 & 0.17 & 0.31 & 0.36 \\
FPFH \cite{rusu09} & 0.62 & 0.60 & 0.39 & 0.14 & 0.25 & 0.32 \\
PointNet \cite{qi17} & 0.58 & 0.48 & 0.38  & 0.18 & 0.17 & 0.25 \\
PointNet++ \cite{qi17_plus} & 0.62  & 0.56 & 0.40 & 0.15 & 0.22 & 0.31 \\
JSIS3D \cite{pham19} & 0.74  & 0.73 & 0.63 & 0.28 & 0.29 & 0.36 \\
3D-BoNet \cite{yang19} & 0.75 & 0.72 & \textbf{0.68} & 0.20 & 0.29 & 0.35 \\
LRGNet & 0.75 & 0.74 & 0.67 & 0.25 & 0.41 & 0.43 \\
\vtop{\hbox{\strut LRGNet +}\hbox{\strut local search}} & \textbf{0.76} & \textbf{0.75} & \textbf{0.68} & \textbf{0.34} & \textbf{0.44} & \textbf{0.45} \\ 
\midrule
\multicolumn{7}{l}{\textit{S3DIS (train) $\rightarrow$ ScanNet (test)}} \\
Region growing & 0.62 & 0.60 & 0.44 & 0.17 & 0.23 & 0.30 \\
Rabbani et al. \cite{rabbani06} & 0.64 & 0.62 & 0.49 & 0.13 & 0.24 & 0.32 \\
FPFH \cite{rusu09} & 0.53 & 0.51 & 0.28 & 0.10 & 0.14 & 0.26 \\
PointNet \cite{qi17} & 0.57 & 0.51 & 0.40  & 0.08 & 0.13 & 0.26 \\
PointNet++ \cite{qi17_plus} & 0.63  & 0.57 & 0.47 & 0.15 & 0.21 & 0.32 \\
JSIS3D \cite{pham19} & 0.57  & 0.56 & 0.31 & 0.15 & 0.13 & 0.22 \\
3D-BoNet \cite{yang19} & 0.59 & 0.54 & 0.34 & 0.10 & 0.13 & 0.24 \\
LRGNet & 0.69 & 0.67 & 0.54 & 0.25 & \textbf{0.33} & \textbf{0.39} \\
\vtop{\hbox{\strut LRGNet +}\hbox{\strut local search}} & \textbf{0.69} & \textbf{0.68} & \textbf{0.56} & \textbf{0.31} & \textbf{0.33} & 0.38 \\
    \bottomrule
  \end{tabular}
\end{table} 

The segmentation performance in indoor scenes was evaluated on the S3DIS dataset \cite{armeni16} and the ScanNet dataset \cite{dai17} alternatingly with one dataset used as training data while the other dataset is held out as test data. The first three evaluation metrics used are from the clustering literature, namely normalized mutual information (NMI), adjusted mutual information (AMI), and adjusted rand index (ARI), as defined in \cite{vinh10}. The other three evaluation metrics used are from the object detection literature, namely precision (PRC), recall (RCL), and mean intersection-over-union (mIOU). For the purposes of calculating precision and recall, true positive detections are defined as cases where the IOU between a predicted point cloud segment and a ground truth point cloud segment is greater than 50\% (as established in \cite{armeni16}). Each evaluation metric is calculated per room of the test dataset and averaged across all rooms (note that this setup means that they differ from the metrics used in \cite{pham19}\cite{yang19} which are averaged across classes). The proposed region growing method is compared against several class-agnostic competing methods including (i) conventional region growing which uses the raw features directly with thresholding, (ii) Rabbani et al. which uses smoothness constraints \cite{zhan09}, (iii) segmentation using the Fast Point Feature Histogram (FPFH) descriptor \cite{rusu09}. The class-specific competing methods are (iv) PointNet \cite{qi17} (v) PointNet++ \cite{qi17_plus} (vi) JSIS3D \cite{pham19} and (vii) 3D-BONET \cite{yang19}. The latter four methods are implemented based on publicly-available open-source code.

Table \ref{table:acc} shows a quantitative comparison of each evaluation metric between the different segmentation methods. Results show that the proposed LRGNet achieves the overall best results across all metrics whether tested on the S3DIS dataset or the ScanNet dataset. In addition, the results also show that applying a local search method (random restart with number of points as the optimality criterion) can further improve the results by optimizing the region growing process of LRGNet. Note that PointNet and PointNet++ do not directly compute the instance labels. Instead, instance labels are obtained by clustering neighboring points that have the same semantic label, as described in \cite{qi17}. This process could cause inaccuracies which lead to worse performance in Table 1 when compared to other methods such as region growing or FPFH. Furthermore, this generalization study also demonstrates the importance of having a class-agnostic point cloud segmentation method. ScanNet is collected in mostly home environments whereas S3DIS is collected in mostly office environments which contain different classes of objects. As a result, most methods that are trained with class information on the S3DIS dataset do not perform as well on the ScanNet dataset. On the other hand, class-agnostic methods such as the proposed LRGNet are able to better generalize to the new point cloud scenes.

\begin{figure*}[t]
  \centering
  \includegraphics[clip,width=0.99\linewidth]{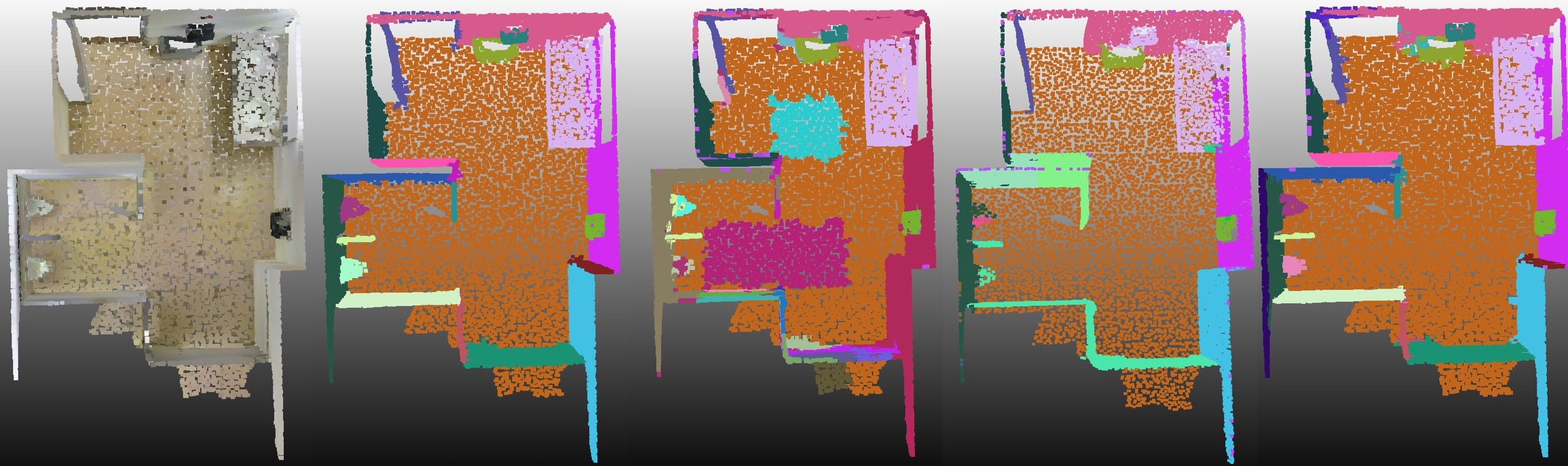}
  \caption{Visualization of segmentation results on the S3DIS dataset: (from left to right) (i) original RGB point cloud (ii) ground truth (iii) PointNet++ (iv) 3D-BoNet (v) proposed method }
  \label{fig:result_viz}
\end{figure*}

\begin{figure*}[t]
  \centering
  \includegraphics[clip,width=0.99\linewidth]{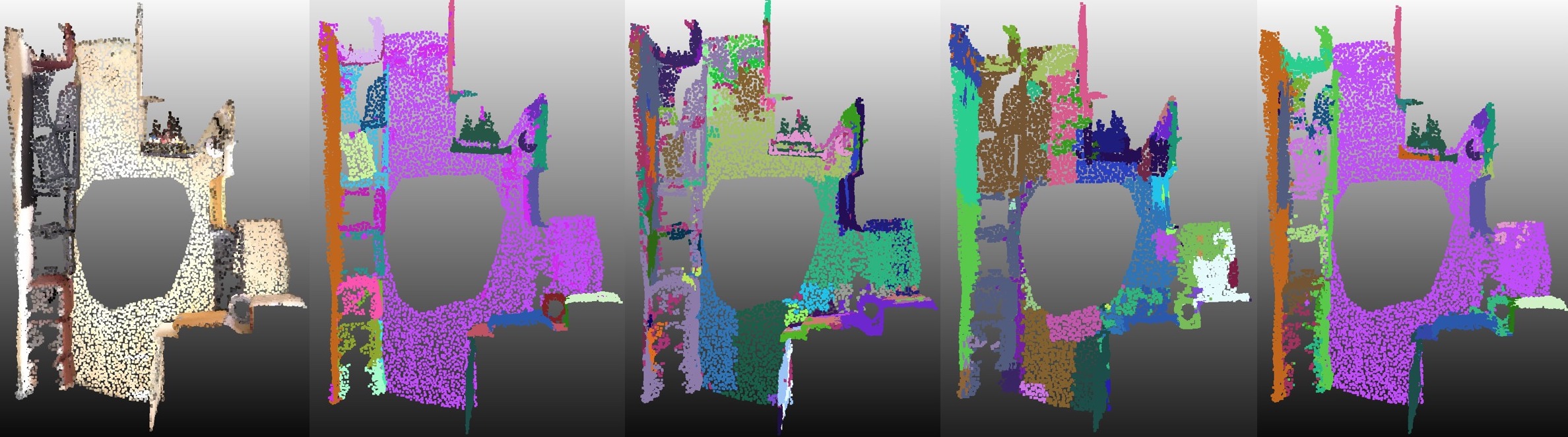}
  \caption{Visualization of segmentation results on the ScanNet dataset: (from left to right) (i) original RGB point cloud (ii) ground truth (iii) PointNet++ (iv) 3D-BoNet (v) proposed method }
  \label{fig:sn_combined}
\end{figure*}

Figures \ref{fig:result_viz} and \ref{fig:sn_combined} show a qualitative comparison of the segmentation results from an overhead perspective, with the ceiling removed to improve visualization. Results show that competing methods have more cases of undersegmentation or oversegmentation, whereas the proposed method is able to perform better since it is directly trained on optimizing the region growing segmentation process.

\subsection{Segmentation in outdoor scenes}

\begin{figure*}[t]
  \centering
  \includegraphics[clip,width=0.99\linewidth]{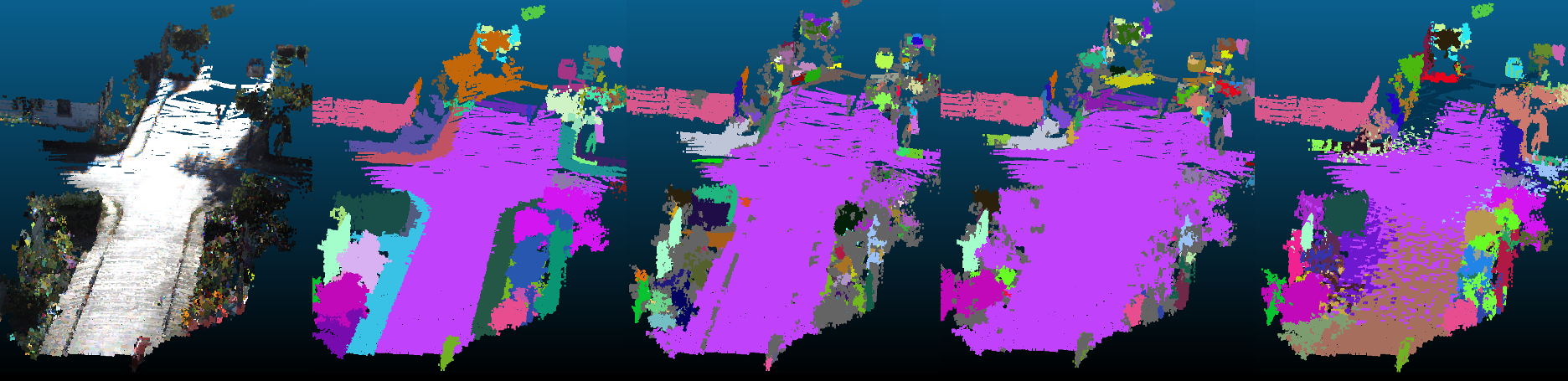}
  \caption{Visualization of segmentation results on the Semantic KITTI dataset: (from left to right) (i) original RGB point cloud (ii) ground truth (iii) simple region growing (iv) FPFH (v) proposed method }
  \label{fig:kitti_combined}
\end{figure*}

The segmentation performance in outdoor scenes is evaluated on the Semantic KITTI dataset \cite{behley2019iccv}. To  be  consistent  with  the  experimental  setup  in  indoor  scenes,  the original  Semantic KITTI dataset  is  modified  as  follows:  (i)  RGB  channels are  projected  from  the  left  camera  images  to  the  LiDAR scans to get color-mapped point clouds, (ii) the semantic labels are discarded and only instance labels are used during training, (iii) points without a valid  RGB  color  and  instance  label  are  filtered  out, (iv) scan data from 20  consecutive  LiDAR  scans  are  merged  to  get  a  dense point cloud. 10 driving sequences are used as training data whereas 1 driving sequence is held out as test data. Figure \ref{fig:kitti_combined} shows a qualitative comparison between the proposed method and other class-agnostic segmentation methods such as simple region growing and FPFH. Results show that the proposed method is able to better separate out the road from sidewalk areas compared to other methods which tend to undersegment those areas. The proposed method is also able to generate more complete segments of building and vegetation compared to other methods which tend to oversegment those objects.

\subsection{Ablation studies}

Table \ref{table:ablation} shows a series of ablation studies of the segmentation performance on the S3DIS dataset for different architecture design choices. In these ablation studies, Areas 1,2,3,4,6 of S3DIS are used as the training set whereas Area 5 of S3DIS is used as the validation set. These ablation studies include using the original method but (i) without using the \textit{remove mask} prediction to remove outlier points from the current region, (ii) without using the minimum curvature criteria when selecting seed points, instead selecting them randomly, (iii) without normalizing the input point features, (iv) only using XYZ features, (v) only using XYZ+RGB features, (vi) using $I=128$, $J=128$, or (vii) using $I=256$, $J=256$. Results show that these variants lead to a 5-10\% change in performance but the complete method has the best performance overall. 

\begin{table}[h]
  \caption{Ablation study of different design choices on S3DIS dataset}
  \label{table:ablation}
  \centering
  \begin{tabular}{ccccccc}
    \toprule
    Method & NMI & AMI & ARI & PRC & RCL & mIOU\\
    \midrule
no remove mask & 0.78 & 0.75 & 0.73 & 0.51 & 0.41 & 0.42 \\
no seed selection & 0.81 & 0.77 & 0.76 & 0.40 & 0.55 & 0.53 \\
\vtop{\hbox{\strut no feature}\hbox{\strut normalization}} & 0.81 & 0.76 & 0.75 & 0.39 & 0.53 & 0.52 \\
\midrule
only XYZ & 0.76 & 0.75 & 0.61 & 0.24 & 0.42 & 0.44 \\ 
only XYZ+RGB & 0.78 & 0.78 & 0.70 & 0.33 & 0.53 & 0.51 \\
\midrule
$I=128$, $J=128$ & 0.79 & 0.79 & 0.69 & 0.32 & 0.57 & 0.55 \\
$I=256$, $J=256$ & 0.80 & 0.80 & 0.74 & 0.35 & 0.56 & 0.54 \\
\midrule
complete method & 0.81 & 0.78 & 0.77 & 0.43 & 0.56 & 0.54 \\
    \bottomrule
  \end{tabular}
\end{table} 

Table \ref{table:local} shows the segmentation performance on the S3DIS dataset when comparing different local search methods based on the quality (NMI, AMI, ARS metrics) as well as efficiency (average number of inference steps needed to segment one object instance). The NMI, AMI, and ARS metrics are calculated as mean $\pm$ standard deviation across all rooms of the S3DIS dataset. Random restart is applied with a total of 10 restarts whereas beam search is applied with 3 candidates each expanded 3 times. This setting is so that random restart and beam search will have roughly the same number of computation steps. Note that the non-greedy methods use a noisier randomized selection criteria compared to the greedy method, so the number of computational steps is not exactly proportional to the number of restarts (or beam width). Overall, both random restart and beam search improved the segmentation performance compared to the greedy method, with \textit{random restart - NP} having the best performance overall. However, this comes with the computational tradeoff of having many more passes through the prediction network for each object instance.

\begin{table}[h]
  \caption{Segmentation performance on S3DIS dataset with different local search methods}
  \label{table:local}
  \centering
  \begin{tabular}{ccccccc}
    \toprule
    Method & NMI(\%) & AMI(\%) & ARS(\%) & Avg steps\\
    \midrule
greedy & 82$\pm$4& 73$\pm$7 & 74$\pm$12 & 13.38 \\
random restart - ML & 82$\pm$5 & 78$\pm$7 & 77$\pm$12 & 190.06 \\
random restart - NP & 82$\pm$4 & 79$\pm$6 & 77$\pm$10 & 188.70 \\
beam search - ML & 82$\pm$5 & 78$\pm$7 & 77$\pm$12 & 159.79 \\
beam search - NP & 82$\pm$4 & 78$\pm$6 & 77$\pm$10 & 175.50 \\
    \bottomrule
  \end{tabular}
\end{table}

\subsection{Computation time analysis}

Table \ref{table:time} shows a comparison of the computation time involved to completely segment a point cloud scene, evaluated on a randomly selected subset of 50 scenes from the S3DIS dataset. The computation time is measured by the authors on an Intel Xeon E3-1200 CPU with a NVIDIA GTX1080 GPU. Compared to other methods, LRGNet lies on the higher end of the spectrum in terms of computation cost due to having to run a large number of region growing iterations to segment each scene. In future work, using parallel processing methods or an adaptive region growing step size could potentially improve the computational speed.

\begin{table}[h]
  \caption{Computation time comparison for segmentation per scene}
  \label{table:time}
  \centering
  \begin{tabular}{cccc}
    \toprule
Method & \multicolumn{3}{c}{Computation time (s)} \\
 & Min & Mean & Max \\
    \midrule
Region growing & 0.4 & 4.8 & 18.6 \\
Rabbani et al. & 0.3 & 4.6 & 18.3 \\
FPFH & 0.5 & 4.6 & 17.8 \\
Pointnet & 0.1 & 0.6 & 2.5 \\
PointNet++ & 0.1 & 0.9 & 3.5 \\
JSIS3D & 1.0 & 539.2 & 16713.9 \\
3D-BoNet & 1.5 & 14.1 & 69.3 \\
LRGNet & 0.8 & 64.9 & 620.9 \\
\bottomrule
  \end{tabular}
\end{table}

\iffalse
\begin{table}[h]
  \caption{Breakdown of computation time involved in each component}
  \label{table:breakdown}
  \centering
  \begin{tabular}{ccc}
    \toprule
Component & Average Time (s) & Percentage (\%) \\
    \midrule
Feature pre-processing & 3.89 $\pm$ 2.74 & 6.0 \\
Network inference & 4.01 $\pm$ 2.75 & 6.2 \\
Neighbor set update & 1.73 $\pm$ 1.97 & 2.7 \\
Inlier set update & 55.29 $\pm$ 88.47 & 85.1 \\
    \midrule
Total & 64.92 & 100.0 \\    
\bottomrule
  \end{tabular}
\end{table}
\fi

\section{CONCLUSIONS}
3D point cloud segmentation is important for robots to be able to understand the layout and geometry of objects in the surrounding environment. To achieve this, this paper proposes a learnable region growing method that can perform class-agnostic point cloud segmentation. The proposed method works by decomposing the segmentation problem into a series of region growing subproblems. A deep neural network is trained to predict how to add or remove points from a point cloud region to morph it into incrementally more complete regions of an object instance. The advantages of the proposed method are that is can be used to segment objects of arbitrary shape, size, or class, and that it is more accurate and generalizable compared to existing methods.


\section*{ACKNOWLEDGMENT}

The work reported herein was supported by the United States Air Force Office of Scientific Research (Award No. FA2386-17-1-4655) and by a grant (18CTAP-C144787-01) funded by the Ministry of Land, Infrastructure, and Transport (MOLIT) of Korea Agency for Infrastructure Technology Advancement (KAIA). Any opinions, findings, and conclusions or recommendations expressed in this material are those of the authors and do not necessarily reflect the views of the United States Air Force and MOLIT. Special thanks to John Yi Seon Keun for assisting with the segmentation experiments.

%%%%%%%%%%%%%%%%%%%%%%%%%%%%%%%%%%%%%%%%%%%%%%%%%%%%%%%%%%%%%%%%%%%%%%%%%%%%%%%%

{\small
\bibliographystyle{IEEEtran}
\bibliography{main}

\begin{thebibliography}{10}
\providecommand{\url}[1]{#1}
\csname url@samestyle\endcsname
\providecommand{\newblock}{\relax}
\providecommand{\bibinfo}[2]{#2}
\providecommand{\BIBentrySTDinterwordspacing}{\spaceskip=0pt\relax}
\providecommand{\BIBentryALTinterwordstretchfactor}{4}
\providecommand{\BIBentryALTinterwordspacing}{\spaceskip=\fontdimen2\font plus
\BIBentryALTinterwordstretchfactor\fontdimen3\font minus
  \fontdimen4\font\relax}
\providecommand{\BIBforeignlanguage}[2]{{%
\expandafter\ifx\csname l@#1\endcsname\relax
\typeout{** WARNING: IEEEtran.bst: No hyphenation pattern has been}%
\typeout{** loaded for the language `#1'. Using the pattern for}%
\typeout{** the default language instead.}%
\else
\language=\csname l@#1\endcsname
\fi
#2}}
\providecommand{\BIBdecl}{\relax}
\BIBdecl

\bibitem{chen18ur}
J.~{Chen}, P.~{Kim}, Y.~K. {Cho}, and J.~{Ueda}, ``Object-sensitive potential
  fields for mobile robot navigation and mapping in indoor environments,'' in
  \emph{2018 15th International Conference on Ubiquitous Robots (UR)}, June
  2018.

\bibitem{dube18}
R.~Dube, M.~G. Gollub, H.~Sommer, I.~Gilitschenski, R.~Siegwart, C.~Cadena, and
  J.~Nieto, ``Incremental-segment-based localization in 3-d point clouds,''
  \emph{IEEE Robotics and Automation Letters}, vol.~3, no.~3, July 2018.

\bibitem{kim17}
P.~Kim, J.~Chen, and Y.~K. Cho, ``Robotic sensing and object recognition from
  thermal-mapped point clouds,'' \emph{International Journal of Intelligent
  Robotics and Applications}, vol.~1, no.~3, Sep 2017.

\bibitem{qi17}
C.~R. Qi, H.~Su, K.~Mo, and L.~J. Guibas, ``Pointnet: Deep learning on point
  sets for 3d classification and segmentation,'' \emph{Proc. Computer Vision
  and Pattern Recognition (CVPR), IEEE}, 2017.

\bibitem{law18}
A.~C.~C. Law, N.~Southon, N.~Senin, P.~Stavroulakis, R.~Leach, R.~D. Goodridge,
  and Z.~Kong, ``Curvature-based segmentation of powder bed point clouds for
  inprocess monitoring,'' in \emph{Proceedings of the 29th Annual International
  Solid Freeform Fabrication Symposium}, 2018.

\bibitem{rusu10}
R.~B. Rusu, ``Semantic 3d object maps for everyday manipulation in human living
  environments,'' \emph{KI - Künstliche Intelligenz}, vol.~24, no.~4, 2010.

\bibitem{wang2017}
X.~Wang, L.~Zou, X.~Shen, Y.~Ren, and Y.~Qin, ``A region-growing approach for
  automatic outcrop fracture extraction from a three-dimensional point cloud,''
  \emph{Computers and Geosciences}, vol.~99, 2017.

\bibitem{wangweiyue18}
W.~Wang, R.~Yu, Q.~Huang, and U.~Neumann, ``Sgpn: Similarity group proposal
  network for 3d point cloud instance segmentation,'' in \emph{The IEEE
  Conference on Computer Vision and Pattern Recognition (CVPR)}, June 2018.

\bibitem{pham19}
Q.-H. Pham, T.~Nguyen, B.-S. Hua, G.~Roig, and S.-K. Yeung, ``Jsis3d: Joint
  semantic-instance segmentation of 3d point clouds with multi-task pointwise
  networks and multi-value conditional random fields,'' in \emph{The IEEE
  Conference on Computer Vision and Pattern Recognition (CVPR)}, June 2019.

\bibitem{yang19}
B.~Yang, J.~Wang, R.~Clark, Q.~Hu, S.~Wang, A.~Markham, and N.~Trigoni,
  ``Learning object bounding boxes for 3d instance segmentation on point
  clouds,'' in \emph{Advances in Neural Information Processing Systems 32},
  2019.

\bibitem{armeni16}
I.~Armeni, O.~Sener, A.~R. Zamir, H.~Jiang, I.~Brilakis, M.~Fischer, and
  S.~Savarese, ``3d semantic parsing of large-scale indoor spaces,'' in
  \emph{Proceedings of the IEEE International Conference on Computer Vision and
  Pattern Recognition}, 2016.

\bibitem{dai17}
A.~Dai, A.~X. Chang, M.~Savva, M.~Halber, T.~Funkhouser, and M.~Nie{\ss}ner,
  ``Scannet: Richly-annotated 3d reconstructions of indoor scenes,'' in
  \emph{Proc. Computer Vision and Pattern Recognition (CVPR), IEEE}, 2017.

\bibitem{sahba06}
F.~{Sahba}, H.~R. {Tizhoosh}, and M.~M.~A. {Salama}, ``A reinforcement learning
  framework for medical image segmentation,'' in \emph{The 2006 IEEE
  International Joint Conference on Neural Network Proceedings}, July 2006.

\bibitem{bhardwaj17}
M.~Bhardwaj, S.~Choudhury, and S.~A. Scherer, ``Learning heuristic search via
  imitation,'' in \emph{1st Annual Conference on Robot Learning, CoRL 2017,
  Mountain View, California, USA, November 13-15, 2017, Proceedings}, ser.
  Proceedings of Machine Learning Research, vol.~78.\hskip 1em plus 0.5em minus
  0.4em\relax {PMLR}, 2017.

\bibitem{derivaux10}
S.~Derivaux, G.~Forestier, C.~Wemmert, and S.~Lefèvre, ``Supervised image
  segmentation using watershed transform, fuzzy classification and evolutionary
  computation,'' \emph{Pattern Recognition Letters}, vol.~31, no.~15, 2010.

\bibitem{zhu14}
X.~X. {Zhu} and M.~{Shahzad}, ``Facade reconstruction using multiview
  spaceborne tomosar point clouds,'' \emph{IEEE Transactions on Geoscience and
  Remote Sensing}, vol.~52, no.~6, June 2014.

\bibitem{ahmed20}
S.~M. Ahmed and C.~M. Chew, ``Density-based clustering for 3d object detection
  in point clouds,'' in \emph{Proceedings of the IEEE/CVF Conference on
  Computer Vision and Pattern Recognition (CVPR)}, June 2020.

\bibitem{xie20}
C.~Xie, Y.~Xiang, A.~Mousavian, and D.~Fox, ``Unseen object instance
  segmentation for robotic environments,'' in \emph{arXiv:2007.08073}, 2020.

\bibitem{hu20}
P.~{Hu}, D.~{Held}, and D.~{Ramanan}, ``Learning to optimally segment point
  clouds,'' \emph{IEEE Robotics and Automation Letters}, vol.~5, no.~2, April
  2020.

\bibitem{mv3d}
X.~Chen, H.~Ma, J.~Wan, B.~Li, and T.~Xia, ``Multi-view 3d object detection
  network for autonomous driving,'' in \emph{IEEE CVPR}, 2017.

\bibitem{lang19}
A.~H. Lang, S.~Vora, H.~Caesar, L.~Zhou, J.~Yang, and O.~Beijbom,
  ``Pointpillars: Fast encoders for object detection from point clouds,'' in
  \emph{Proceedings of the IEEE/CVF Conference on Computer Vision and Pattern
  Recognition (CVPR)}, June 2019.

\bibitem{shi19}
S.~Shi, X.~Wang, and H.~Li, ``Pointrcnn: 3d object proposal generation and
  detection from point cloud,'' in \emph{The IEEE Conference on Computer Vision
  and Pattern Recognition (CVPR)}, June 2019.

\bibitem{liu17}
F.~Liu, S.~Li, L.~Zhang, C.~Zhou, R.~Ye, Y.~Wang, and J.~Lu, ``3dcnn-dqn-rnn: A
  deep reinforcement learning framework for semantic parsing of large-scale 3d
  point clouds,'' in \emph{The IEEE International Conference on Computer Vision
  (ICCV)}, Oct 2017.

\bibitem{qi17_plus}
C.~R. Qi, L.~Yi, H.~Su, and L.~J. Guibas, ``Pointnet++: Deep hierarchical
  feature learning on point sets in a metric space,'' \emph{arXiv preprint
  arXiv:1706.02413}, 2017.

\bibitem{jianghaiyong20}
H.~Jiang, F.~Yan, J.~Cai, J.~Zheng, and J.~Xiao, ``End-to-end 3d point cloud
  instance segmentation without detection,'' in \emph{Proceedings of the
  IEEE/CVF Conference on Computer Vision and Pattern Recognition (CVPR)}, June
  2020.

\bibitem{jiang20}
L.~Jiang, H.~Zhao, S.~Shi, S.~Liu, C.-W. Fu, and J.~Jia, ``Pointgroup: Dual-set
  point grouping for 3d instance segmentation,'' \emph{Proceedings of the IEEE
  Conference on Computer Vision and Pattern Recognition (CVPR)}, 2020.

\bibitem{engelmann20}
F.~Engelmann, M.~Bokeloh, A.~Fathi, B.~Leibe, and M.~Niessner, ``3d-mpa:
  Multi-proposal aggregation for 3d semantic instance segmentation,'' in
  \emph{Proceedings of the IEEE/CVF Conference on Computer Vision and Pattern
  Recognition (CVPR)}, June 2020.

\bibitem{he20}
T.~H. He, Y.~Liu, C.~Shen, W.~Xinlong, and C.~Sun, ``Instance-aware embedding
  for point cloud instance segmentation,'' in \emph{Proceedings of the European
  Conference on Computer Vision (ECCV)}, 2020.

\bibitem{dimitrov15}
A.~Dimitrov and M.~Golparvar-Fard, ``Segmentation of building point cloud
  models including detailed architectural/structural features and mep
  systems,'' \emph{Automation in Construction}, vol.~51, 2015.

\bibitem{chen19jcce}
J.~Chen, Z.~Kira, and Y.~K. Cho, ``Deep learning approach to point cloud scene
  understanding for automated scan to 3d reconstruction,'' \emph{Journal of
  Computing in Civil Engineering}, vol.~33, no.~4, 2019.

\bibitem{umbach94}
R.~{Haeb-Umbach} and H.~{Ney}, ``Improvements in beam search for 10000-word
  continuous-speech recognition,'' \emph{IEEE Transactions on Speech and Audio
  Processing}, vol.~2, no.~2, April 1994.

\bibitem{rabbani06}
T.~Rabbani, F.~{van den Heuvel}, and G.~Vosselman,
  ``\BIBforeignlanguage{English}{Segmentation of point clouds using smoothness
  constraints},'' in \emph{\BIBforeignlanguage{English}{ISPRS 2006 :
  Proceedings of the ISPRS commission V symposium Vol. 35, part 6 : image
  engineering and vision metrology, Dresden, Germany 25-27 September 2006}},
  H.~Maas and D.~Schneider, Eds., vol.~35.\hskip 1em plus 0.5em minus
  0.4em\relax International Society for Photogrammetry and Remote Sensing
  (ISPRS), 2006.

\bibitem{rusu09}
R.~B. {Rusu}, N.~{Blodow}, and M.~{Beetz}, ``Fast point feature histograms
  (fpfh) for 3d registration,'' in \emph{2009 IEEE International Conference on
  Robotics and Automation}, 2009.

\bibitem{vinh10}
N.~X. Vinh, J.~Epps, and J.~Bailey, ``Information theoretic measures for
  clusterings comparison: Variants, properties, normalization and correction
  for chance,'' \emph{J. Mach. Learn. Res.}, vol.~11, Dec. 2010.

\bibitem{zhan09}
Q.~Zhan, Y.~Liang, and Y.~Xiao, ``Color-based segmentation of point clouds,''
  in \emph{IAPRS XXXVIII Part 3/W8}, 2009.

\bibitem{behley2019iccv}
J.~Behley, M.~Garbade, A.~Milioto, J.~Quenzel, S.~Behnke, C.~Stachniss, and
  J.~Gall, ``{SemanticKITTI: A Dataset for Semantic Scene Understanding of
  LiDAR Sequences},'' in \emph{Proc. of the IEEE/CVF International Conf.~on
  Computer Vision (ICCV)}, 2019.

\end{thebibliography}
}

\end{document}